\documentclass[11pt]{article}

\usepackage[preprint]{acl}

\usepackage[T1]{fontenc}

\usepackage{microtype}
\usepackage{inconsolata}

\usepackage{tipa}
\usepackage{pifont}

\newcommand{\bng}[1]{%
  {\fontfamily{bang}\selectfont #1}%
}

\def\bng{\bngx}

\font\bngx=bang10

\def\*#1*#2{o\null{#2}{#1}}

\def\sh#1{\setbox0=\hbox{#1}%
     \kern-.02em\copy0\kern-\wd0
     \kern.04em\copy0\kern-\wd0
     \kern-.02em\raise.0433em\box0 }

\usepackage{graphicx}
\usepackage{caption}
\usepackage{adjustbox}

\usepackage{multirow}
\usepackage{booktabs}
\usepackage{amsmath,mathtools}
\usepackage{array}
\usepackage{listings}
\usepackage{cleveref}
\usepackage{pdfpages}
\usepackage{soulutf8} 
\usepackage{tikz}

\definecolor{pink}{RGB}{250,162,184}

\newcommand{\colorboxsquare}[2][pink]{%
  \tikz[baseline=(b.base)]{
    \node (b) [inner sep=0pt] {};
    \fill[#1] (0,0) rectangle (#2,#2);
  }%
}

\definecolor{purplekeywords}{rgb}{0.5, 0, 0.5}

\title{\textsc{Reveal}-Bangla: A Dataset \\ for Cross-Lingual Multi-Step Reasoning Evaluation}

\author{Khondoker Ittehadul Islam ~\;~~\;~~\;~ Gabriele Sarti \\[5pt]
  Center for Language and Cognition (CLCG), University of Groningen \\[5pt]
  \texttt{k.i.islam@student.rug.nl} ~\;~ \texttt{g.sarti@rug.nl}
}

\begin{document}
\maketitle
\begin{abstract}
Language models have demonstrated remarkable performance on complex multi-step reasoning tasks. However, their evaluation has been predominantly confined to high-resource languages such as English. In this paper, we introduce a manually translated Bangla multi-step reasoning dataset derived from the English \textsc{Reveal} dataset, featuring both binary and non-binary question types. We conduct a controlled evaluation of English-centric and Bangla-centric multilingual small language models on the original dataset and our translated version to compare their ability to exploit relevant reasoning steps to produce correct answers. Our results show that, in comparable settings, reasoning context is beneficial for more challenging non-binary questions, but models struggle to employ relevant Bangla reasoning steps effectively. We conclude by exploring how reasoning steps contribute to models' predictions, highlighting different trends across models and languages.\footnote{Dataset:\hspace{0.3em}\texttt{\url{https://huggingface.co/datasets/khondoker/reveal-bangla}}, licensed CC-BY-ND 4.0}
\end{abstract}

\section{Introduction}

\begin{figure}[t]
    \centering
    \includegraphics[width=\linewidth]{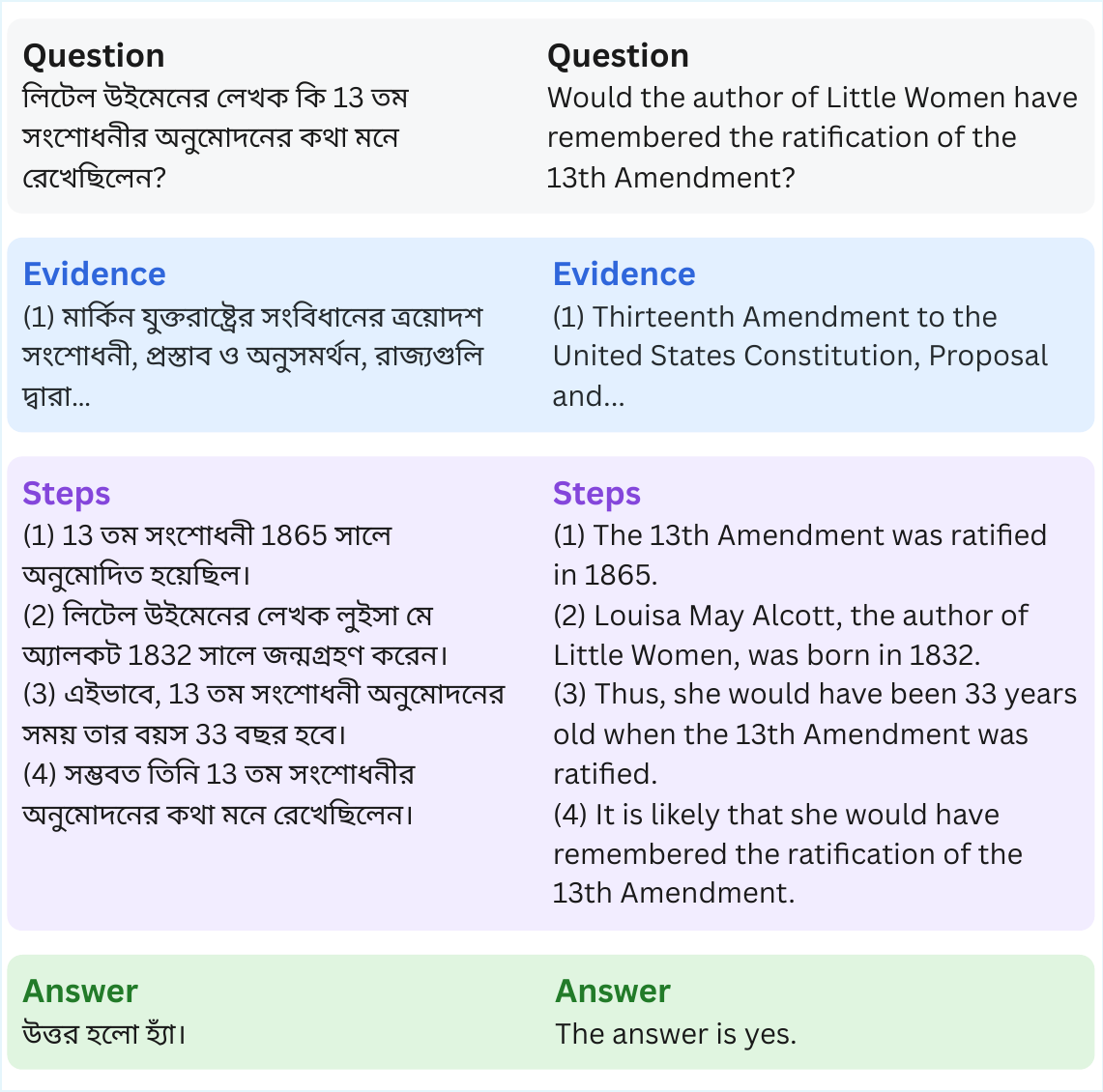}
    \caption{A Row instance of \textbf{\textsc{Reveal}-Bangla} containing translated Question, Evidence, Reasoning Steps and Answer from \textsc{Reveal}.}
    \label{fig:sample_row}
    \vspace{-10pt}
\end{figure}

Large Language Models (LLMs) have demonstrated remarkable versatility across a wide spectrum of natural language processing tasks \citep{radford2019language}. A pivotal breakthrough in enhancing their complex reasoning capabilities has been the introduction of Chain-of-Thought (CoT) prompting \citep{wei2022chain}, which encourages models to generate intermediate reasoning steps before arriving at final answers, yielding substantial performance improvements \citep{white2024livebench, wang2022self}.
Despite these advances, the evaluation of LLM reasoning capabilities remains heavily skewed toward high-resource languages, creating significant gaps in our understanding of how these models perform across linguistically diverse contexts. In this work, we focus specifically on the Bangla language, which boasts 268 million speakers and ranks as the sixth most spoken language globally\footnote{\url{https://en.wikipedia.org/wiki/List_of_languages_by_total_number_of_speakers}}, particularly for its computationally challenging
morphological richness \cite{choudhury2007evolution, das2010automatic}. As the native language of Bangladesh and the second most prominent Indo-Aryan language after Hindi \cite{ethnologue_2021}, Bangla represents a critical case study for cross-lingual reasoning evaluation.
The growing technological transformation in densely populated and economically emerging regions where Bangla is spoken \cite{bangladesh-tech-strategy} underscores the urgent need for developing faithful AI technologies that can enhance social welfare and economic opportunities. While recent work in Bangla focused on simple extractive or multiple-choice question answering \citet{ekram-etal-2022-banglarqa, shafayat-etal-2024-benqa, rony2024banglaquadbengaliopendomainquestion}, to our knowledge, no datasets with human-validated reasoning steps are available for this language. This lack of resources hinders our ability to assess and improve the reasoning capabilities of LLMs in the Bangla language.

In this work, we address this gap by introducing \textbf{\textsc{Reveal}-Bangla}, a manually translated Bangla version of a subset of the English \textsc{Reveal} dataset, containing annotated multi-step reasoning chains with gold answers. We exploit our resource and its original English counterpart to evaluate the abilities of two small language models---both proficient in Bangla and English, but one predominantly English-centric, and the other mainly Bangla-centric---in exploiting reasoning step to produce the correct answers given a query, following recent work showing how non-English languages can harm reasoning abilities in LLMs \citep{qi2025modelsreasonlanguagecontrolling}.

Moreover, recent cross-lingual studies have revealed that generated reasoning chains often exhibit inconsistencies and produce misleading intermediate steps, raising questions about their explanatory reliability \citep{lanham2023measuringfaithfulnesschainofthoughtreasoning,paul-etal-2024-making}. To address these concerns, post-hoc attribution techniques have emerged as valuable tools for analyzing models' internal processes by assigning importance scores to context elements such as retrieved documents \citep{qi-sarti-etal-2024-model, cohen2024contextcite} and reasoning chains \citep{cohenwang2025learning}, thereby revealing their contribution to final predictions. We exploit a similar methodology using the \textsc{ContextCite} method~\citep{cohen2024contextcite} to examine how reasoning steps contribute to model answers in English and Bangla, highlighting different patterns of importance across the two languages. 

\section{Related Work}

Large Language Models (LLMs) operate as probabilistic sequence predictors, estimating the likelihood of the next token given previous context \cite{vaswani2017attention, radford2019language}. For practical application, explicit training on instructions was found to further improve answer quality~\citep{sanh2022multitask,wang2022self}. Recently, eliciting reasoning from LLMs, e.g. via step-by-step Chain of Thought reasoning~(CoT, \citealp{wei2022chain}), was found to further improve the response accuracy for complex queries.

Some popular reasoning datasets in English include \textsc{StrategyQA} \citep{geva2021did}, featuring reasoning- and knowledge-intensive yes/no queries; \textsc{Fermi} \citep{kalyan2021much}, comprising estimation questions that require numerical answers and a blend of knowledge and reasoning; \textsc{MuSiQue} \cite{trivedi2022musique}, which includes multi-hop reasoning questions with free-text entity answers, generated from Wikipedia paragraphs; and \textsc{Sports Understanding} \cite{srivastava2022beyond}, consisting of yes/no questions that demand reasoning about sports players, leagues, and maneuvers. \citet{jacovi2024chain} combined the aforementioned datasets and human-annotated each LLM-generated step in terms of \textbf{attribution} relative to provided Wikipedia paragraphs and \textbf{logical coherence} in light of previous reasoning steps. The resulting dataset, dubbed \textsc{Reveal}, was used to prompt capable LLMs in a chain-of-thought setting and analyzed using Natural Language Inference (NLI) classifiers to evaluate model-generated responses. In our work, we manually translate a subset of \textsc{Reveal} into Bangla and adopt their setup to evaluate cross-lingual English-Bangla models.

Recently, \citet{jin2024impact} explored small and large parameterized models, revealing a linear relationship between accuracy and the number of reasoning steps. We conduct a similar analysis, focusing particularly on small language models (SLMs) with a manageable size (1B parameters). SLMs were recently found capable of high-quality answers in RAG setups~\citep{huang-etal-2024-less, small-rag-cs}, with relevant input information compensating for their limited reasoning abilities. In this work, we use annotated CoTs produced by larger LLMs to investigate whether SLMs can effectively leverage reasoning information in English and Bangla.

\section{Development of \textsc{Reveal}-Bangla }
\subsection{Data Collection}
\label{sec:data-collection}
We start by selecting a subset of the \textsc{Reveal} dataset.\footnote{\url{https://huggingface.co/datasets/google/reveal}} Provided we want to test the ability of SLMs to obtain the correct answer given valid reasoning chains, we focus specifically on examples having all reasoning steps as either logical or fully attributable to the provided Wikipedia paragraphs. Furthermore, among the three models considered by the \textsc{Reveal} authors to generate answers, we decided to choose the two models with the most answers, i.e., Flan-UL2-20B \cite{tay2022ul2} and GPT-3 (text-davinci-003, \citealp{brown2020language}).\footnote{We do not include Flan-PaLM-540B \cite{longpre2023flan} due to our limited evaluation resources.}  

We obtain a total of $104$ unique questions, with $188$ evidence paragraphs and $355$ reasoning steps. While only $60\%$ of all the steps are fully attributed to context, all steps are logically relevant. The dataset contains $70\%$ yes--no \textit{binary} questions, making it especially fitting for verifying the relevance of reasoning steps towards a simple atomic answer.\footnote{We present the counts and tokens distribution of steps and evidence in Figures \ref{fig:step_cnt_token_combined_figures} and \ref{fig:evidence_cnt_token_combined_figures} in the Appendix.}


\paragraph{Translation} The English$\to$Bangla translation of the selected subset (751 texts) was performed by a native Bangla-speaking graduate student. During the translation process, some digits and certain terms were left unchanged, for instance \textit{76ers} (a basketball team in the NBA), \textit{g/dL} (Grams per decilitre), /\textipa{kæv@ndIS}/ (pronunciation of Henry Cavendish), \textit{Équipe d'Haïti de football} (French spelling of the Haitian National Football Team), \textit{inter alia}. As an additional analysis, we assess the quality of automatic translations from Google Translate on the same subset, finding high-quality outputs for health and historical data, but subpar performance on the \textsc{Sports Understanding} subset (examples in Appendix \ref{sec:google_error_eg}). Generally, automatic translations were of higher quality when performed one sentence at a time. We employ only the manually translated subset in our evaluation.

\section{Evaluation}

\paragraph{Model Selection} For our evaluation on \textsc{Reveal}  and our Bangla variant, we use Llama-3.2-1B-Instruct (or \textit{EngLlama}) \cite{grattafiori2024llama} and BanglaLLama-3.2-1b-bangla-alpaca-orca-instruct-v0.0.1 (or \textit{BenLlama}) \cite{zehady2024bongllama}. \textit{EngLlama} is a popular English-centric multilingual SLM, while \textit{BenLlama} is a Bangla-centric model fine-tuned from \textit{EngLlama} using \textsc{Bangla-Alpaca-Orca}, a collection of instruction tuning examples including the popular \textsc{Alpaca} and \textsc{OpenOrca} datasets \citep{alpaca,openorca} automatically translated into Bangla. Importantly, despite their different language focus, both models maintain answering capabilities in both English and Bangla, motivating our cross-lingual analysis.

\paragraph{Prompting Setup} We experiment our methods on two main settings: (1) \texttt{gen\_ans}, where the model produces an answer without any reasoning step and (2) \texttt{w\_cot\_gen\_ans}, where we provide the model with the annotated reasoning steps from \textsc{Reveal} and \textsc{Reveal}-Bangla.\footnote{A third setting prompting the model to generate its own reasoning steps, \texttt{gen\_cot\_ans}, was not included due to the poor performance of SLMs on CoT reasoning.} Both models were tested on the English and Bangla \textsc{Reveal} subsets containing the same examples, using a prompt including the query and relevant evidence paragraphs, plus the reasoning steps in the \texttt{w\_cot\_gen\_ans} setting.\footnote{Examples of prompt templates for each setting are available in Appendix \ref{sec:chat_eg}.} We test our models on Nvidia A100 GPU, using greedy decoding for reproducible results, and limiting output length to 256 tokens. Reasoning steps in the \texttt{w\_cot\_gen\_ans} setting are appended to the assistant portion of the chat, using \texttt{continue\_final\_message = True} to let the model complete the generation by producing a final answer. We leave the remaining generation parameters unchanged.

\paragraph{Verifiers} To verify the accuracy of the model-generated final answer against the actual final answer, we choose mDeBERTa-v3-base-xnli-multilingual-nli-2mil7 \cite{laurer_less_2022} model as it is the only NLI model that supports both English and Bangla. We consider \textit{entailment} labels as correct answers and \textit{contradict} as otherwise. As this NLI model additionally verdicts \textit{neutral}, authors manually verify the response to classify it as valid or not. Furthermore, as language detection tools such as \texttt{langdetect} \cite{nakatani2010langdetect} do not support Bangla, we manually assign \textit{contradict} to answers generated in scripts that do not match English or Bangla in the respective settings. We provide our \textit{hypothesis} and \textit{premise} NLI template in the Appendix \ref{sec:nli_strct_eg}. We also present additional limitations of the multi-lingual NLI model in the Appendix \ref{sec:nli_limit_eg}, to foster research on cross-lingual NLI comprising Bangla.

\begin{figure}[t]
    \centering
    \includegraphics[width=0.95\linewidth]{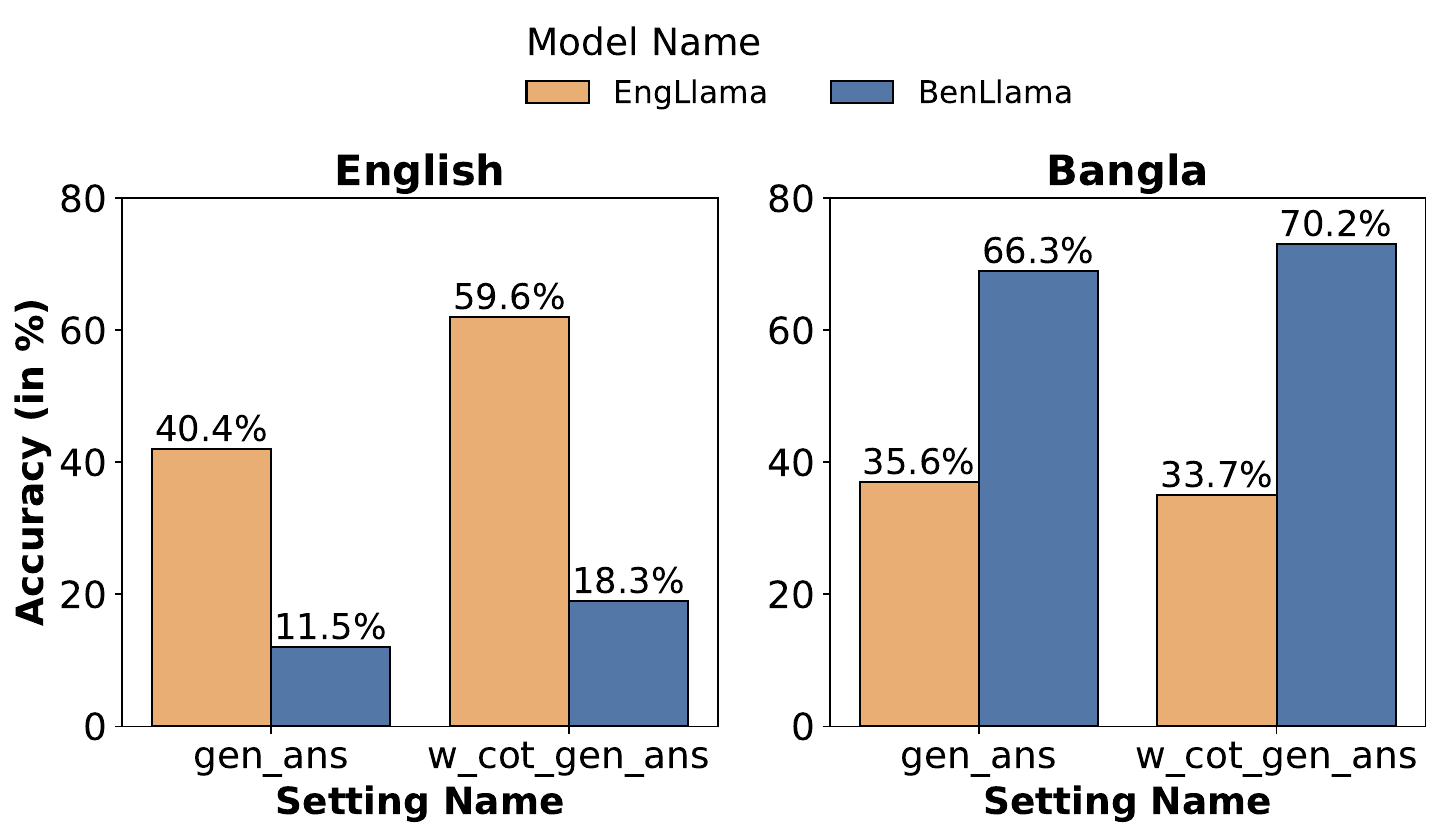}
    \caption{Accuracy of \textit{EngLlama} and \textit{BenLlama} for the \texttt{gen\_ans} and \texttt{w\_cot\_gen\_ans} settings on English and Bangla \textsc{Reveal} subsets.}
    \label{fig:overall_accuracy}
\end{figure}

\paragraph{Results} Figure \ref{fig:overall_accuracy} shows the accuracy of tested models in both languages. Unsurprisingly, we find both models performed better on their respective main languages. Moreover, despite their small size, both models were generally found to effectively use the provided reasoning steps to further improve their accuracy. However, we observe that \textit{EngLlama} obtains worse performances when given \texttt{w\_cot\_gen\_ans} steps in Bangla (35.6\% $\to$ 33.7\%), and find that CoT gains for the \textit{BenLlama} model in Bangla are much milder than for the \textit{EngLlama} model in English (+3.9\% vs. +19.2\%). These results confirm that, in the less-resourced Bangla setting, \textit{additional relevant reasoning information may not be sufficient to mitigate the limited language capabilities of the tested SLMs}, especially when a Bangla-specific tuning was not performed, as was the case for \textit{EngLlama}.

\begin{figure}[t]
    \centering
    \includegraphics[width=0.95\linewidth]{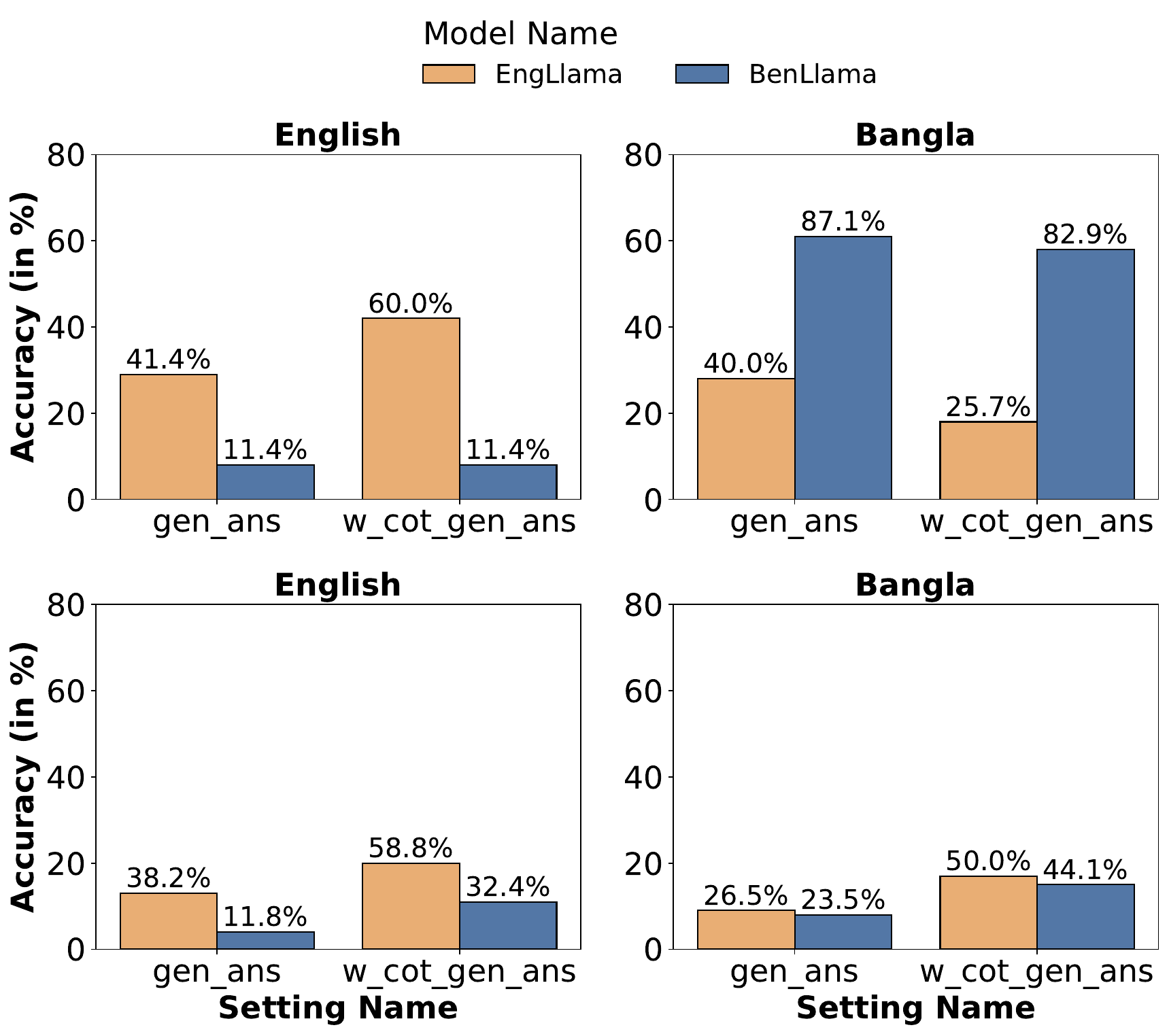}
    \caption{\textit{EngLlama} and \textit{BenLlama} accuracy on \textsc{Reveal} Binary (top) and Non-Binary (bottom) questions.}
    \label{fig:all_qtype_performance}
    \vspace{-10pt}
\end{figure}

We further examine model performances across on binary and non-binary questions in the selected \textsc{Reveal} subset in Figure \ref{fig:all_qtype_performance}. We find that the \textit{EngLlama} model excels in non-binary questions across both languages, outperforming the \textit{BenLlama} model in both \texttt{gen\_ans} and \texttt{w\_cot\_gen\_ans} settings, even in Bangla by a narrow margin. The stronger performance of \textit{BenLlama} in the aggregate case is largely motivated by binary questions, in which the model obtains accuracy $>80\%$. We also find that while CoT steps have an uneven effect on binary questions, they are consistently beneficial for non-binary ones, across both models and languages. This confirms previous findings on the limited effectiveness of CoT in simpler settings by \citet{liu2024mind}, and suggests the benefits of CoT generalize even to less-resourced languages.

\begin{figure}[t]
    \centering
    \includegraphics[width=0.95\linewidth]{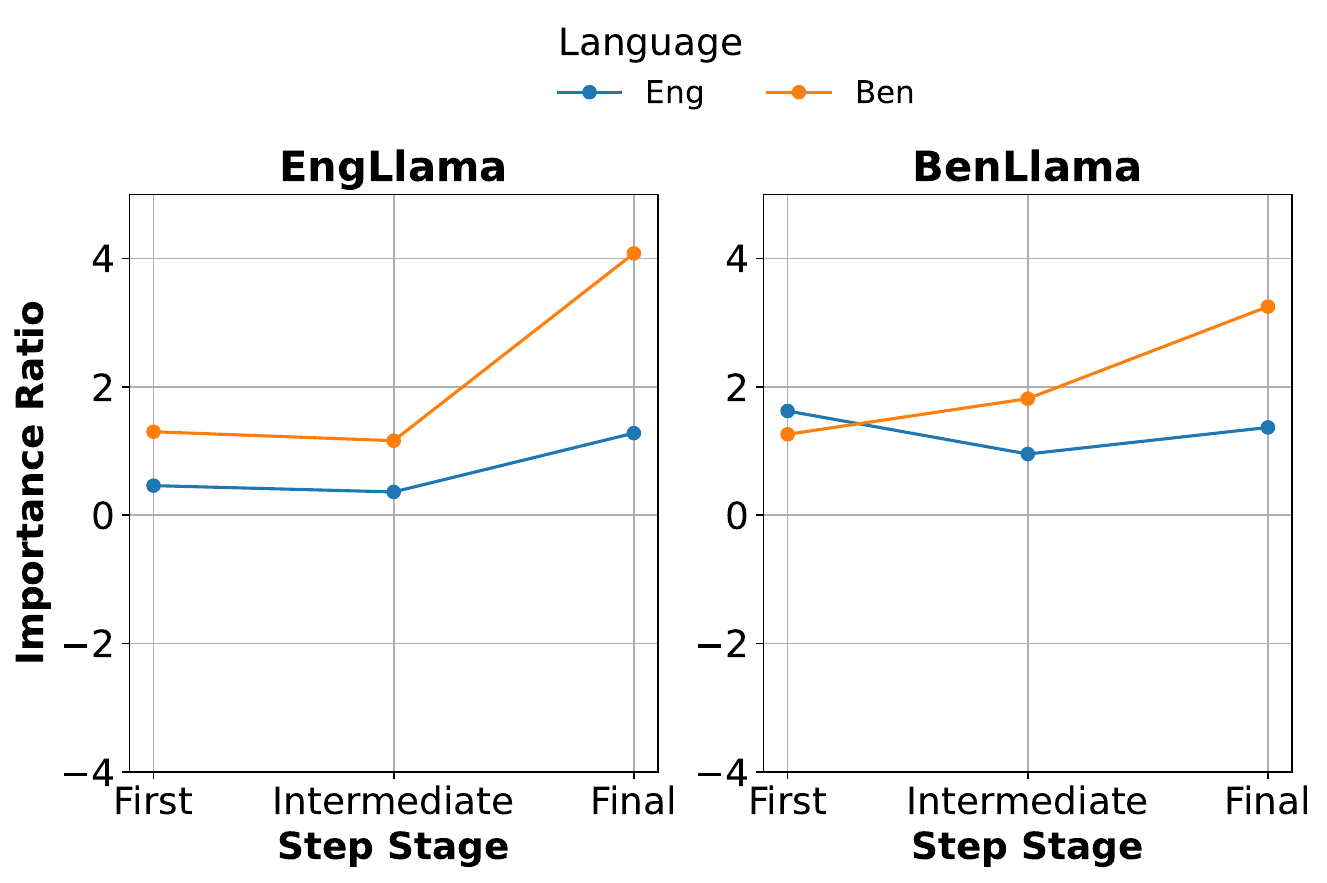}
    \caption{Importance ratio for \textit{EngLlama} and \textit{BenLlama} on \texttt{w\_cot\_gen\_ans} reasoning steps between $-4$ (lowest) and $+4$ (highest).
}
    \label{fig:context_cite_w_cot_gen_ans_perform_eval}
    \vspace{-10pt}
\end{figure}

\paragraph{Attributing Answers to Reasoning Steps} To conclude our analysis, we conduct a preliminary investigation into how CoT steps influence model answers. We employ \texttt{ContextCite} \cite{cohen2024contextcite} to attribute the final answer generated by the model to the provided reasoning steps in the \texttt{w\_cot\_gen\_ans} setting using surrogate linear models, an approach similar to LIME~\citep{ribeiro-etal-2016-trust}. Figure \ref{fig:context_cite_w_cot_gen_ans_perform_eval} presents an overview of our results for the two models across both languages. We observe that in most cases, later steps tend to have a larger influence on the model response. 
This suggests that the models place higher emphasis on answer-specific information located in later steps more than on understanding the context provided in earlier steps. This highlights the inherent limitations of these models in context comprehension, which is essential for answering complex questions. Future research could investigate whether this trend holds with larger model sizes.
Additionally, we find that both models accord high importance to the Bangla language. We speculate that Bangla's morphological richness causes to assign larger values across attention layers.
We leave the exploration of model interpretability in low-resource languages as an interesting direction for future work.

\section{Conclusion}
We presented \textsc{Reveal}-Bangla, a manually translated portion of the popular English multi-step reasoning dataset \textsc{Reveal}. Our cross-lingual analysis of SLMs revealed limited performance gains from CoT reasoning in the less-resourced Bangla setting compared to English, with gains primarily involving more complex non-binary questions. Further investigation into attributing reasoning steps highlighted differences in importance across models and languages. These findings underscore the need for developing language-specific approaches to enhance reasoning capabilities in low-resource languages, rather than directly transferring techniques optimized for English.

\section*{Limitations}

\paragraph{Dataset Scale and Coverage} Our study is constrained by the relatively small scale of the translated dataset, comprising only 104 unique questions from the original REVEAL dataset. This limited sample size may not fully capture the diversity of reasoning patterns and linguistic phenomena present in Bangla. Additionally, the 70\% skew toward binary questions may not accurately reflect real-world reasoning scenarios, potentially overestimating model performance on more complex, open-ended reasoning tasks.

\paragraph{Model Selection Constraints} We restricted our evaluation to small language models with 1B parameters due to computational constraints. While this choice enables insights into resource-efficient deployment scenarios, it limits our understanding of how larger, more capable models might leverage Bangla reasoning steps. The exclusion of the \texttt{gen\_cot\_ans} setting, where models generate their own reasoning chains, further restricts our analysis to scenarios with gold reasoning steps, which may not reflect realistic deployment conditions.

\paragraph{Translation and Annotation Quality} Although we employed manual translation by a native Bangla speaker, the translation was performed by a single annotator without inter-annotator agreement measures. This approach may introduce individual biases or inconsistencies in translation choices, particularly for domain-specific terminology in sports and medical contexts. The preservation of certain English terms and pronunciations, while necessary, may also affect how models process the hybrid text.

\paragraph{Evaluation Methodology Limitations} Our reliance on the mDeBERTa-v3-base-xnli model for answer verification introduces its own limitations, as acknowledged in our appendix. The model's tendency to produce neutral verdicts required manual intervention, potentially introducing subjective judgments. Furthermore, the absence of Bangla-specific language detection tools necessitated manual script verification, which may not scale to larger evaluations.

\paragraph{Cross-lingual Generalization} Our findings are specific to the English-Bangla language pair and may not generalize to other low-resource languages with different linguistic properties, writing systems, or relationships to English. The choice of \textit{BenLlama}, which was fine-tuned on automatically translated instruction data, may also introduce artifacts from machine translation that affect our conclusions about Bangla reasoning capabilities.

\paragraph{Attribution Analysis Scope} Our investigation into reasoning step attribution using ContextCite represents only a preliminary analysis. The surrogate linear model approach may not capture complex non-linear interactions between reasoning steps, and we did not explore alternative attribution methods that might reveal different patterns of step importance across languages.

\bibliography{custom}

\onecolumn
\appendix

\section*{Appendix}
\section{Dataset}
\subsection{Sample}
\begin{table}[ht!]
\footnotesize
\centering
\renewcommand{\arraystretch}{1.40}
 \begin{tabular}{ l m{0.80\columnwidth}  } 
 
 \hline
\textbf{Question [E]} & Can a Bengal cat survive eating only pancakes? \\
 \midrule
 
\textbf{Question [B]} & \bng{ebNG/gl ibDal ik shudhu pYanekk ekhJe ebNNec thaket paer?} \\
  \midrule
 
\textbf{Evidence [E]} &  1. Carnivore, Obligate carnivores: Obligate carnivores are diverse. The amphibian axolotl consumes mainly worms and larvae in its environment, but if necessary will consume algae. All felids, including the domestic cat, require a diet of primarily animal flesh and organs. Specifically, cats have high protein requirements and their metabolisms appear unable to synthesize essential nutrients such as retinol, arginine, taurine, and arachidonic acid; thus, in nature, they must consume flesh to supply these nutrients. \\
\vspace{0.5em} 
& 2. Pancake: A pancake (or hotcake, griddlecake, or flapjack) is a flat cake, often thin and round, prepared from a starch-based batter that may contain eggs, milk and butter and cooked on a hot surface such as a griddle or frying pan, often frying with oil or butter. Archaeological evidence suggests that pancakes were probably the earliest and most widespread cereal food eaten in prehistoric societies. \\
  \midrule
 
\textbf{Evidence [B]} &  \bng{1. maNNGsashii, badhY maNNGsashii: badhY maNNGsashii {oi}bictRYmJ. Ubhcr AYaek/salTl tar pirebesh pRdhant krRim EbNNG lar/bha khaJ, teb pRJeajen esh{O}la gRas kreb. grRHpailt ibDal sH sms/t ekKetRr jnY pRathimkbhaeb pshur maNNGs EbNNG ANG/gguilr EkiT khadY pRJeajn. ibeshSht, ibDaledr Uc/c epRaiTenr pRJeajniiJta thaek EbNNG taedr ibpak eriTnl, Aarijna{I}n, TaUirn EbNNG AYaraikeDaink AYaiseDr meta pRJeajniiJ puiSh/T sNNGeshLShN kret AkKm bel men HJ; E{I}bhaeb, pRkrRitet, E{I} puiSh/T srbraH krar jnY taedr AbshY{I} maNNGs gRHN kret Heb.
} \\
\vspace{0.5em} 
& 2. \bng{pYanekk: EkiT pYanekk (ba HTekk, igRDl ekk ba phLYapjYak) EkiT phLYaT ekk, pRaJsh{I} patla EbNNG egalakar HJ, Ja EkiT s/Tar/c-ibhit/tk bYaTar ethek {oi}tir kra HJ Jaet iDm, dudh EbNNG makhn thaket paer EbNNG EkiT grm prReSh/Th ran/na kra HJ eJmn EkiT bhaja ba phRa{I}NNG pYan, pRaJsh{I} etl ba makhn idJe bhaja HJ. pRt/ntait/tWk pRmaNguil ethek jana JaJ eJ pYanekkguil sm/bhbt pRa{oi}gitHaisk smaej kha{O}Ja sbecJe pRaciin EbNNG sr/baidhk ibs/trRt khadYshsY ichl.} \\
  \midrule
 
\textbf{Steps [E]} &  1. Cats are obligate carnivores, meaning they need to eat meat to survive. \\
\vspace{0.5em} 
& 2. Pancakes are not a source of meat. \\
\vspace{0.5em} 
& 3. Thus, a Bengal cat cannot survive eating only pancakes. \\

  \midrule

\textbf{Steps [B]} &  \bng{1. ibDal badhYtamuulk maNNGsashii, Jar Ar/th taedr ebNNec thakar jnY maNNGs ekhet Heb.
} \\ 
\vspace{0.5em} 
& \bng{2. pYanekk maNNGesr Ut//s nJ.} \\
\vspace{0.5em} 
& \bng{3. sutraNNG, EkiT ebNG/gl ibDal shudhumatR pYanekk ekhJe bNNacet paer na.} \\
  \midrule
 
\textbf{Answer [E]} & The answer is no. \\
  \midrule
 
\textbf{Answer [B]} & \bng{Ut/tr Hela na.} \\
    \bottomrule
    
 
 
\end{tabular}
\caption{Samples from our dataset comprising of question, evidence, steps, and answer where \textbf{[E]} and \textbf{[B]} following them represents corresponding English and Bangla versions respectively. }
\label{tab:sample_dataset}
\end{table}

\twocolumn

%

\clearpage

\onecolumn
\subsection{Step Count and Token Distribution}
\begin{figure*}[h]
    \centering
    \includegraphics[width=0.95\linewidth]{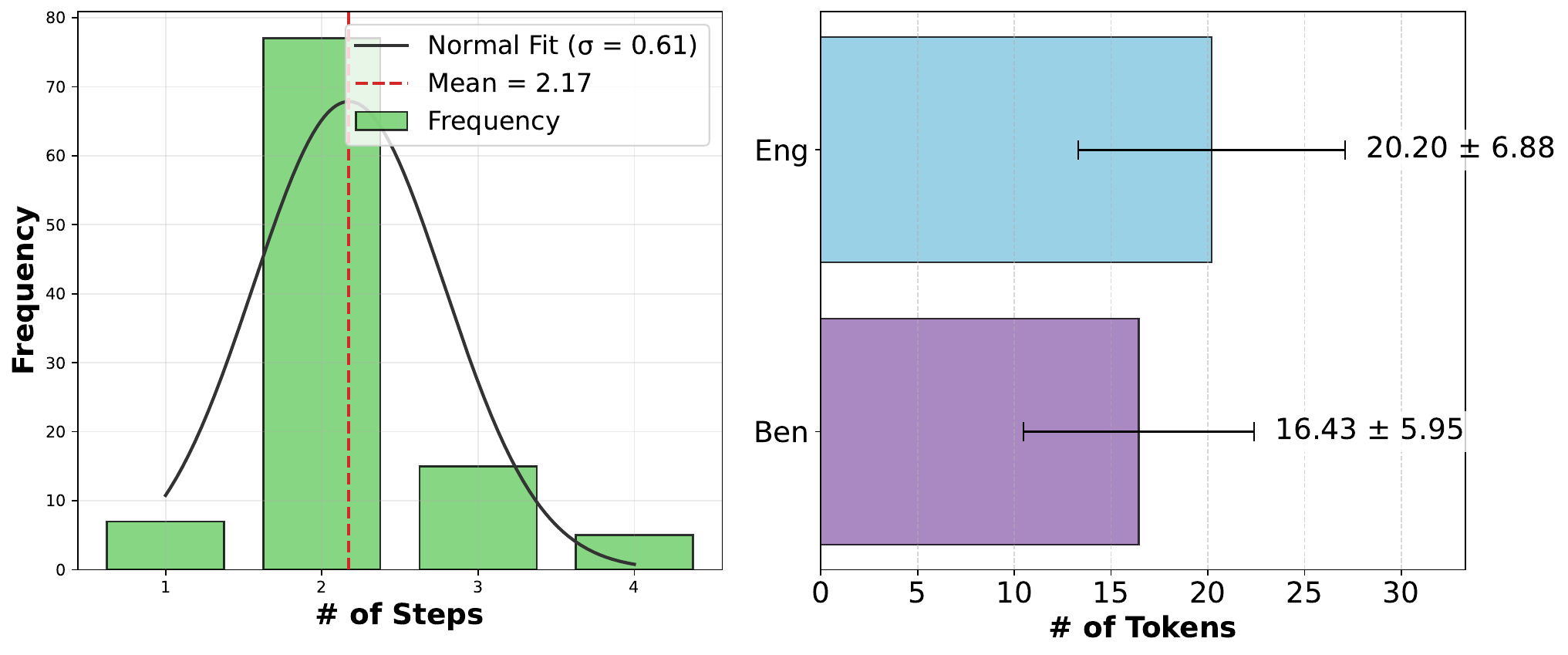}
    \caption{Distribution of Step Count and Token Distribution of Steps. Furthermore, interestingly, number of words required to describe a step in Bangla is less than of English.}
    \label{fig:step_cnt_token_combined_figures}
\end{figure*}

\subsection{Evidence Count and Token Distribution}

\begin{figure*}[h!]
    \centering
    \includegraphics[width=0.95\linewidth]{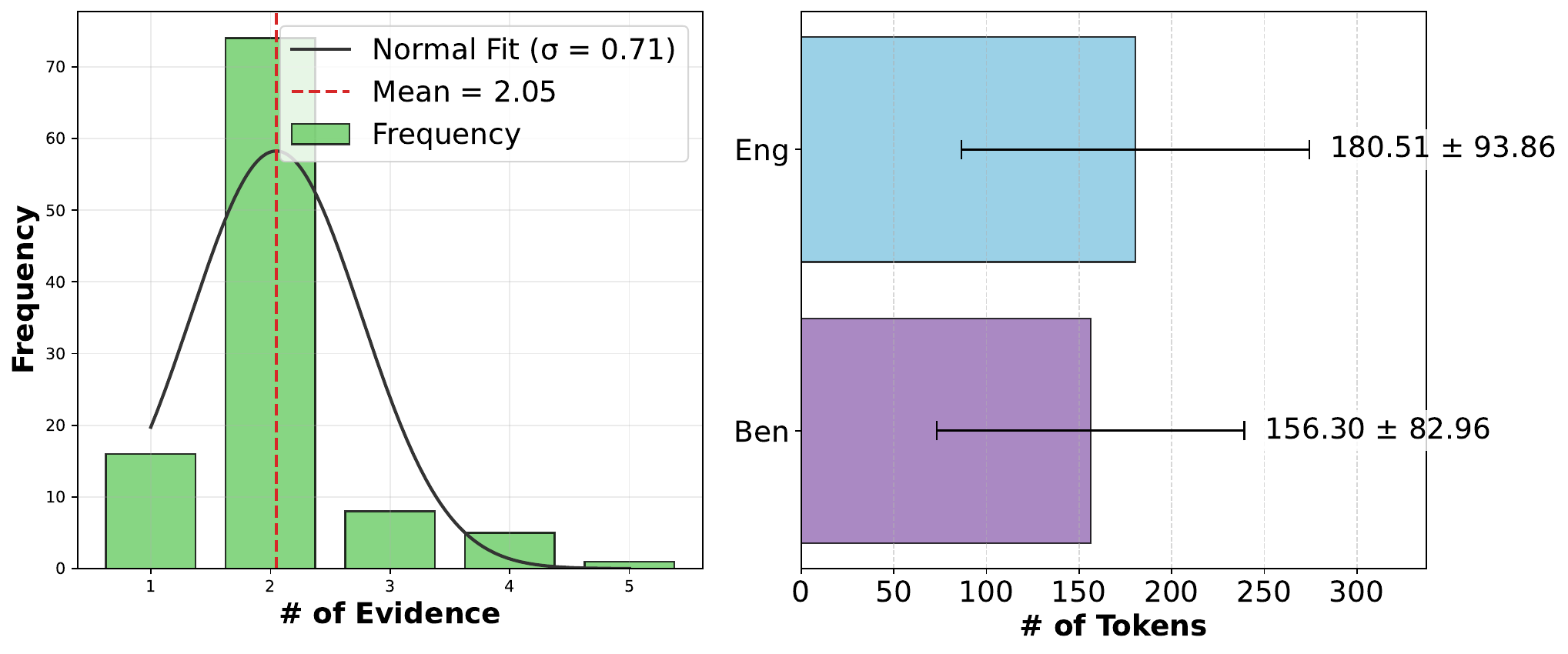}
    \caption{Distribution of Evidence Count and Token Distribution of Steps. On average, there were three evidences associated alongside the questions.}
    \label{fig:evidence_cnt_token_combined_figures}
\end{figure*}
\clearpage

\onecolumn
\section{Example of Chat Prompt Templates}
\label{sec:chat_eg}
\subsection{Setting: \texttt{gen\_ans}}
\begin{table}[ht!]
\centering
\footnotesize
\renewcommand{\arraystretch}{1.40}
 \begin{tabular}{ l m{0.80\columnwidth}  } 

 \hline
\textbf{En} & <|begin\_of\_text|><|start\_header\_id|>system<|end\_header\_id|> \\
 \vspace{0.5em} 
 & You are a helpful assistant. Your goal is to respond to user queries using the provided evidence paragraphs. The final line must contain the word 'Answer:' followed by the answer to the user query.
The response should contain ONLY the final response. If the question requires a yes/no answer, answer using only "yes" or "no". Do NOT provide any additional explanation or comments.<|eot\_id|><|start\_header\_id|>user<|end\_header\_id|> \\
\vspace{0.5em} 
& \# Evidence \\
& 1. Toilet paper, Description, Materials: Toilet paper is usually manufactured from pulpwood trees, but is also sometimes made from sugar cane byproducts or bamboo. \\
& 2. Logging: Logging is the process of cutting, processing, and moving trees to a location for transport. It may include skidding, on-site processing, and loading of trees or logs onto trucks or skeleton cars. \\
\vspace{0.5em} 
& \# Question: \\
&  Would it be hard to get toilet paper if there were no loggers?<|eot\_id|>\\&<|start\_header\_id|>assistant<|end\_header\_id|> \\
\vspace{0.5em} 
& Answer: \\
 \midrule
 
\textbf{Bn} & <|begin\_of\_text|><|start\_header\_id|>system<|end\_header\_id|>\\
\vspace{0.5em} 
& {\bng Aapin Ekjn Upkarii sHkarii. Aapnar Ued/dshY Hl pRdt/t pRmaN Anuec/chd bYbHar ker bYbHarkariir pResh/nr Ut/tr ed{O}Ja. cuuDan/t la{I}en AbshY{I} 'Ut/tr:' shb/diT thakeb EbNNG tarper bYbHarkariir pResh/nr Ut/tr thakeb.
pRitikRJaiTet shudhumatR cuurhan/t Ut/tr thakeb. pRsh/niTr Jid HNNYa/na Ut/terr pRJeajn HJ, taHel shudhumatR "HNNYa" ba "na" bYbHar ker Ut/tr idn. ekan Aitirk/t bYakhYa ba mn/tbY pRdan krebn na.}<|eot\_id|><|start\_header\_id|>user<|end\_header\_id|>\\
\vspace{0.5em} 
& {\bng \# pRmaN} \\
& 1. {\bng TJelT eppar, br/Nna, UpkrN: TJelT eppar sadharNt pal/pUD gach ethek {oi}tir kra HJ, teb kkhn{O} kkhn{O} Aaekhr Upjat ba bNNash ethek{O} {oi}tir kra HJ.}\\
& 2. {\bng ligNNG: ligNNG Hl pirbHenr jnY gach kaTa, pRikRJakrN EbNNG s/thanan/tr krar pRikRJa. Eet is/kiDNNG, An-sa{I}T pRikRJakrN EbNNG TRak ba Ees/kelTn gaDiet gach ba lg elaD kra An/tr/bhuk/t thaket paer.}\\
\vspace{0.5em} 
& {\bng \# pRsh/n}: \\
& {\bng kaThuir na thakel ik TJelT eppar pa{O}Ja kiThn Heb?}<|eot\_id|>\\&<|start\_header\_id|>assistant<|end\_header\_id|>\\
\vspace{0.5em} 
& {\bng Ut/tr}:\\
 
    \bottomrule
 
\end{tabular}
\caption{An example of a chat prompt template from \texttt{gen\_ans} setting. \textbf{En} is of the corresponding English language and \textbf{Bn} is of the Bangla language.}
\label{tab:chat_eg_gen_ans}
\end{table}



\clearpage

\onecolumn
\subsection{Setting: \texttt{w\_cot\_gen\_ans}}
\begin{table}[ht!]
\centering
\footnotesize
\renewcommand{\arraystretch}{1.40}
 \begin{tabular}{ l m{0.80\columnwidth}  } 

 \hline
\textbf{En} & <|begin\_of\_text|><|start\_header\_id|>system<|end\_header\_id|> \\
 \vspace{-0.4em} 
 & You are a helpful assistant. Your goal is to respond to user queries using the provided evidence paragraphs. The final line must contain the word 'Answer:' followed by the answer to the user query.
The response should contain ONLY the final response. If the question requires a yes/no answer, answer using only "yes" or "no". Do NOT provide any additional explanation or comments.<|eot\_id|><|start\_header\_id|>user<|end\_header\_id|> \\
\vspace{-0.4em} 
& \# Evidence \\
& 1. Toilet paper, Description, Materials: Toilet paper is usually manufactured from pulpwood trees, but is also sometimes made from sugar cane byproducts or bamboo. \\
& 2. Logging: Logging is the process of cutting, processing, and moving trees to a location for transport. It may include skidding, on-site processing, and loading of trees or logs onto trucks or skeleton cars. \\
\vspace{-0.4em} 
& \# Question: \\
&  Would it be hard to get toilet paper if there were no loggers?<|eot\_id|>\\&<|start\_header\_id|>assistant<|end\_header\_id|> \\
\vspace{-0.4em} 
& 1. Toilet paper is made from trees. \\
& 2. Loggers are responsible for cutting down trees. \\
& 3. Thus, without loggers, it would be difficult to get toilet paper. \\
\vspace{-0.4em} 
& Answer: \\
 \midrule
 
\textbf{Bn} & <|begin\_of\_text|><|start\_header\_id|>system<|end\_header\_id|>\\
\vspace{-0.4em} 
& {\bng Aapin Ekjn Upkarii sHkarii. Aapnar Ued/dshY Hl pRdt/t pRmaN Anuec/chd bYbHar ker bYbHarkariir pResh/nr Ut/tr ed{O}Ja. cuuDan/t Ut/tr ed{O}Jar Aaeg, dhaep dhaep Juik/t idebn, pRitiT Juik/tr dhapek EkiT ntun la{I}en sNNGkhYaJuk/t bhaeb tailkabhuk/t krebn. cuuDan/t la{I}en AbshY{I} 'Ut/tr:' shb/diT thakeb EbNNG tarper bYbHarkariir pResh/nr Ut/tr thakeb.
pRitikRJaiTet shudhumatR sNNGkhYaJuk/t Juik/tr dhap EbNNG cuurhan/t Ut/tr thakeb. pRsh/niTr Jid HNNYa/na Ut/terr pRJeajn HJ, taHel shudhumatR "HNNYa" ba "na" bYbHar ker Ut/tr idn. ekan Aitirk/t bYakhYa ba mn/tbY pRdan krebn na.}<|eot\_id|><|start\_header\_id|>user<|end\_header\_id|>\\
\vspace{-0.4em} 
& {\bng \# pRmaN} \\
& 1. {\bng TJelT eppar, br/Nna, UpkrN: TJelT eppar sadharNt pal/pUD gach ethek {oi}tir kra HJ, teb kkhn{O} kkhn{O} Aaekhr Upjat ba bNNash ethek{O} {oi}tir kra HJ.}\\
& 2. {\bng ligNNG: ligNNG Hl pirbHenr jnY gach kaTa, pRikRJakrN EbNNG s/thanan/tr krar pRikRJa. Eet is/kiDNNG, An-sa{I}T pRikRJakrN EbNNG TRak ba Ees/kelTn gaDiet gach ba lg elaD kra An/tr/bhuk/t thaket paer.}\\
\vspace{-0.4em} 
& {\bng \# pRsh/n}: \\
& {\bng kaThuir na thakel ik TJelT eppar pa{O}Ja kiThn Heb?}<|eot\_id|>\\
&<|start\_header\_id|>assistant<|end\_header\_id|>\\
\vspace{-0.4em} 
& 1. {\bng TJelT eppar {oi}tir HJ gach ethek.} \\
& 2. {\bng kaThuirra gach kaTar jnY daJii.} \\
& 3. {\bng E{I}bhaeb, kaThuir chaDa, TJelT eppar pa{O}Ja kiThn Heb.} \\
\vspace{-0.4em} 
& {\bng Ut/tr}:\\
 
 \bottomrule
 
\end{tabular}
\caption{An example of a chat prompt template from \texttt{w\_cot\_gen\_ans} setting. \textbf{En} is of the corresponding English language and \textbf{Bn} is of the Bangla language.}
\label{tab:chat_eg_w_cot_gen_ans}
\end{table}

\clearpage

\twocolumn
\section{NLI}
\subsection{Structure Example}
\label{sec:nli_strct_eg}
\subsubsection{English}
\begin{table}[ht]
    \centering
    \small
    \renewcommand{\arraystretch}{1.40}
     \begin{tabular}{ m{15em} } 
     \hline
     
       \textbf{Hypothesis}: Who does the actress that played mary poppins in the 1964 film play in princess diaries? The answer is Queen Clarisse Renaldi.  \\
       \midrule
     \textbf{Premise}: Who does the actress that played mary poppins in the 1964 film play in princess diaries? The answer is Julie Andrews. \\

     \bottomrule
     
    \end{tabular}
    \caption{Example of the \textbf{Hypothesis} and \textbf{Premise} structure for English language. Here, \textbf{Hypothesis} incorporates ground answer and \textbf{Premise} incorporates model predicted answer.}
    \label{tab:nli_strct_eg_en}
\end{table}

\subsubsection{Bangla}
\begin{table}[ht]
    \centering
    \small
    \renewcommand{\arraystretch}{1.40}
     \begin{tabular}{ m{15em} } 
     \hline
     
       \textbf{Hypothesis}: \bng{inUTn sr/bjniin mHakr/Sh dhRubekr man pirmap krar jnY pdar/thibg/Yaniir kaejr ekKtRiT kii? sutraNNG  Ut/tr Hela pdar/thibd EbNNG rsaJnibd.}  \\
       \midrule
     \textbf{Premise}: \bng{inUTn sr/bjniin mHakr/Sh dhRubekr man pirmap krar jnY pdar/thibg/Yaniir kaejr ekKtRiT kii? Ut/tr Hela inUTn sr/bjniin mHakr/Sh dhRubekr man pirmap krar jnY pdar/thibg/Yaniir kaejr ekKtRiT Hela pRmaN.} \\

     \bottomrule
     
    \end{tabular}
    \caption{Example of the \textbf{Hypothesis} and \textbf{Premise} structure for Bangla language.}
    \label{tab:nli_strct_eg_bn}
\end{table}

\subsection{Example Cases of Limitations on Bangla}
\label{sec:nli_limit_eg}
\subsubsection{Entails Proper Noun Spelling Mistakes}
\begin{table}[ht!]
    \centering
    \small
    \renewcommand{\arraystretch}{1.40}
     \begin{tabular}{ l m{15em} } 
     \hline
     
       \textbf{Hypothesis} & \bng{erDkYap Aibhenta sdesYr pt/nii ek? Ut/tr Hela ishla HYankk.}  \\
       \midrule
         \textbf{Premise} & \bng{erDkYap Aibhenta sdesYr pt/nii ek? Ut/tr Hela ishla HYank} \\
     \midrule
      \textbf{Ground Label} & Contradiction  \\
      \midrule
    \textbf{Predicted Label} & Entailment \\

     \bottomrule
     
    \end{tabular}
    \caption{Example of NLI model incorrectly predicting Entailment where the Premise differs with Hypothesis through a spelling mistake on proper noun {\bng ishla HYankk}.}
    \label{tab:nli_limit_eg_1}
\end{table}

\newpage

\subsubsection{Labels \textit{contradict} on model’s elaborate correct answers on \texttt{binary} question}
\begin{table}[ht]
    \centering
    \small
    \renewcommand{\arraystretch}{1.40}
     \begin{tabular}{ l m{15em} } 
     \hline
     
       \textbf{Hypothesis} & \bng{DeraithJa {O}Jein/DNNG ik epar/shar Ud/bhb jaJga ethek Eesechn? Ut/tr Hela HNNYa.}  \\
       \midrule
     \textbf{Premise} & \bng{DeraithJa {O}Jein/DNNG ik epar/shar Ud/bhb jaJga ethek Eesechn? Ut/tr Hela DeraithJa {O}Jein/DNNG epar/shar Ud/bhb jaJga ethek Eesechn.} \\
     \midrule
      \textbf{Ground Label} & Entailment \\
      \midrule
    \textbf{Predicted Label} & Contradict \\

     \bottomrule
     
    \end{tabular}
    \caption{Example of NLI model incorrectly judging Contradict where the Premise contained elaboration on the single word {\bng HNNYa} ("yes") answer.}
    \label{tab:nli_limit_eg_2}
\end{table}

\subsubsection{Labels \textit{entailment} where the main final answer is missing}
\begin{table}[ht]
    \centering
    \small
    \renewcommand{\arraystretch}{1.40}
     \begin{tabular}{ l m{15em} } 
     \hline
     
       \textbf{Hypothesis} & \bng{mHmMd Aat/tar gaDii kii, eJ ekam/painr DYaTsan {oi}tir kra HJeech, Er nijr? sutraNNG  Ut/tr Hela insan AaliTma.}  \\
       \midrule
     \textbf{Premise} & \bng{mHmMd Aat/tar gaDii kii, eJ ekam/painr DYaTsan {oi}tir kra HJeech, Er nijr? Ut/tr Hela insan emaTr ekaNNG, ilimeTD (eHpbar/n: insan ijedasha kabuishik ga{I}sha) Hl EkiT japanii bHujaitk AeTaemaba{I}l pRs/tutkark Jar sdr dp/tr inish-ku, {I}JeaekaHama, japaen. ekam/painiT insan, {I}niphiniT EbNNG DYaTsun} \\
     \midrule
      \textbf{Ground Label} & Contradict \\
      \midrule
    \textbf{Predicted Label} & Entailment \\

     \bottomrule
     
    \end{tabular}
    \caption{Example of NLI model falsely predicts Entailment where the actual proper noun answer "{\bng insan AaliTma}" (Nissan Altima) is not present in the Premise. }
    \label{tab:nli_limit_eg_3}
\end{table}

\newpage
\section{Google Translation Errors}
\label{sec:google_error_eg}
\subsection{Example on American Football Context}

\begin{figure}[ht]
    \centering
    \includegraphics[width=\linewidth]{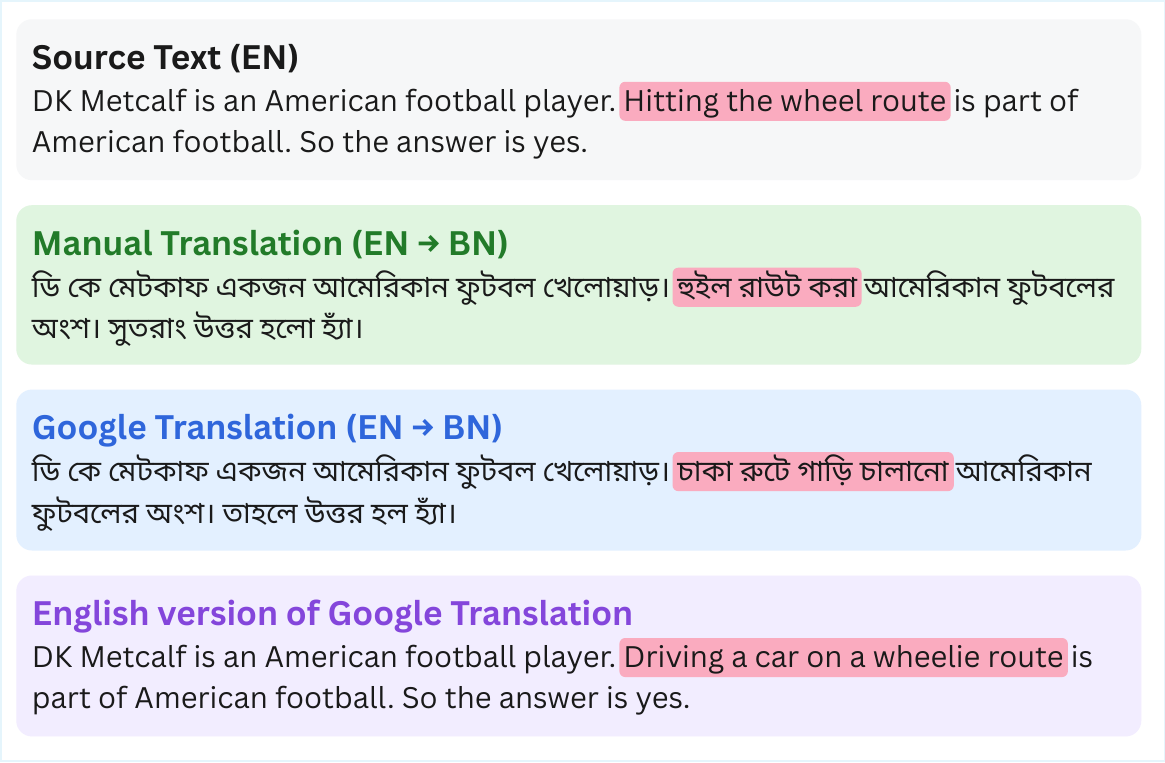}
    \caption{Example of \textbf{Google Translation}'s sub-par performance compared to \textbf{Manual Translation} when given the \textbf{Source} English text with the mistake underlined in pink (\colorboxsquare{0.65em}).}
    \label{fig:google_translation_error_1}
    \vspace{-10pt}
\end{figure}


\subsection{Example on Basketball Context}
\begin{figure}[ht]
    \centering
    \includegraphics[width=\linewidth]{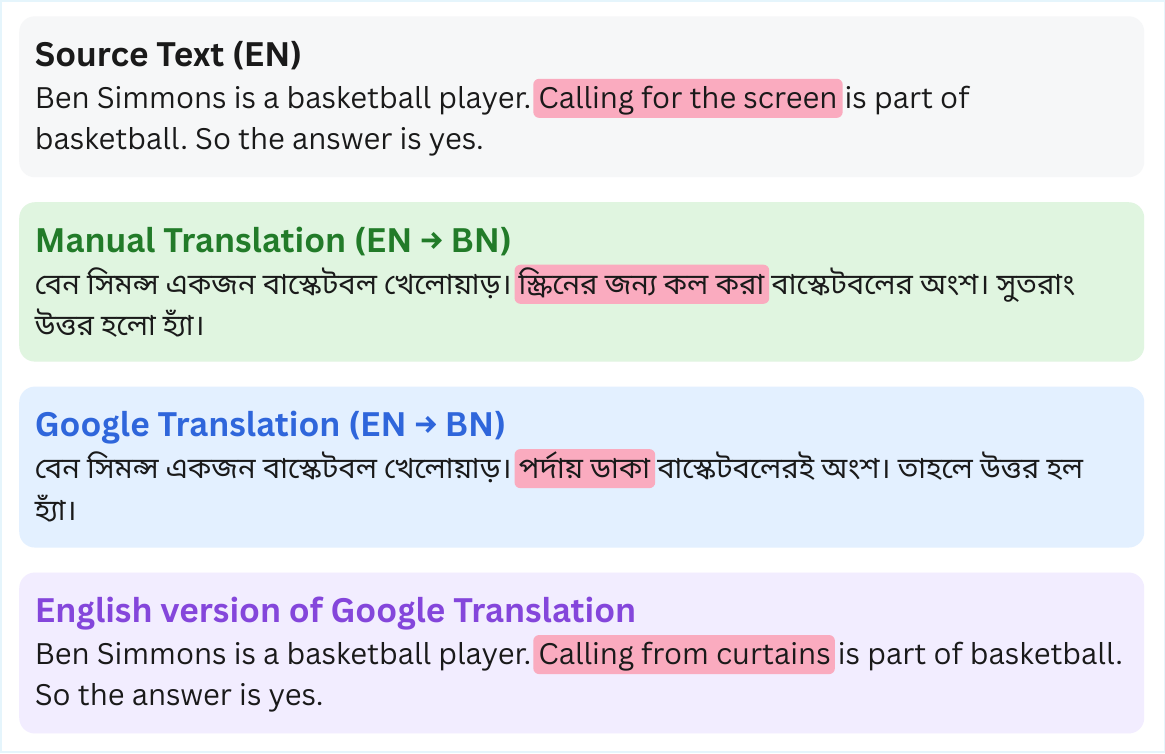}
    \caption{Example of \textbf{Google Translation}'s sub-par performance on Basketball context with the mistake underlined in pink (\colorboxsquare{0.65em}).}
    \label{fig:google_translation_error_2}
    \vspace{-10pt}
\end{figure}

\end{document}